\documentclass[runningheads]{llncs}
\usepackage{graphicx}
\usepackage{amssymb}
\usepackage{amsmath}
\usepackage{multirow}
\usepackage{booktabs}
\usepackage{algorithm}
\usepackage{algorithmic}

\usepackage{enumitem}
\usepackage[misc]{ifsym}
\begin{document}
\title{CGCL: Collaborative Graph Contrastive Learning without Handcrafted Graph Data Augmentations}
\titlerunning{CGCL: Collaborative Graph Contrastive Learning}

%\author{Anonymous Author(s), Paper ID: 1118}
\author{
Tianyu Zhang\inst{1} \and
Yuxiang Ren\inst{2} \inst{(\textrm{\Letter})} \thanks{T. Zhang and Y. Ren---Both authors contribute equally to this paper.} \and
Wenzheng Feng\inst{2} \and
Weitao Du\inst{2,3} \and
Xuecang Zhang\inst{2}
}
%, , 
\authorrunning{T. Zhang et al.}
\institute{
Department of Automation, Tsinghua University, Beijing, China\\
\email{zhangty2016@gmail.com}
\and
Huawei Technologies, China\\
% \email{\{renyuxiang1,fengwenzheng,xxx,xxx,zhangxuecang\}@huawei.com}
% }
\email{\{renyuxiang1, fengwenzheng, zhangxuecang\}@huawei.com}
\and 
Chinese Academy of Sciences\\
\email{duweitao@amss.ac.cn}
}

% \institute{Tsinghua University, Beijing, China \and@
% Huawei Technologies, Germany

%
\maketitle              % typeset the header of the contribution
\begin{abstract}
Unsupervised graph representation learning is a non-trivial topic. The success of contrastive methods in the unsupervised representation learning on structured data inspires similar attempts on the graph. Existing graph contrastive learning (GCL) aims to learn the invariance across multiple augmentation views, which renders it heavily reliant on the handcrafted graph augmentations. However, inappropriate graph data augmentations can potentially jeopardize such invariance. In this paper, we show the potential hazards of inappropriate augmentations and then propose a novel Collaborative Graph Contrastive Learning framework (CGCL). This framework harnesses multiple graph encoders to observe the graph. Features observed from different encoders serve as the contrastive views in contrastive learning, which avoids inducing unstable perturbation and guarantees the invariance. To ensure the collaboration among diverse graph encoders, we propose the concepts of asymmetric architecture and complementary encoders as the design principle. To further prove the rationality, we utilize two quantitative metrics to measure the assembly of CGCL respectively. Extensive experiments demonstrate the advantages of CGCL in unsupervised graph-level representation learning and the potential of collaborative framework. The source code for reproducibility is available at \url{https://github.com/zhangtia16/CGCL}

\keywords{Graph Representation Learning  \and Contrastive Learning \and Collaborative Framework}

\end{abstract}

\section{Introduction}\label{sec:introduction}
% 图重要，图表示学习重要
Graph-structured data contain not only the attribute information of individual units (i.e., nodes) but also the connection information (i.e., edges) between these units. Due to this ability, many applications exhibit the favorable property of graph-structured data, which makes learning effective graph representations for downstream tasks a non-trivial problem.
% GNN+标签稀缺->unsupervised
In recent years, graph neural networks (GNNs)~\cite{velivckovic2017graph,kipf2016semi,sun2019infograph,ying2018hierarchical,ren2020adversarial} have demonstrated excellent performance in graph representation learning. 
%At the node-level, 
 Usually, GNNs learn the graph representation in supervised or semi-supervised scenarios. However, task-specific labels are scarce or unevenly distributed~\cite{zhou2022graph}, which make it time-consuming and labor-intensive to obtain supervised signals. For example, labeling protein data necessitates a large workforce and material resources. 
% 无监督-》自监督-》对比学习-》aug
For the obstacle of scarce labels, unsupervised graph representation learning has emerged as the critical technology to achieve breakthroughs. 
%Graph kernels~\cite{ivanov2018anonymous,shervashidze2011weisfeiler,yanardag2015deep} can learn the representations in an unsupervised manner, but the handcrafted kernel features may lead to poor generalization performance. Self-supervised graph learning methods~\cite{hu2020gpt,jin2020self,you2020does} define pretext tasks as their supervision to implement unsupervised graph representation learning. The pretext tasks are designed based on heuristics and presuppose a specific set of representational invariance (e.g., pairwise attribute similarity~\cite{jin2020self}), whereas downstream tasks may not meet this presupposition (e.g., A classification problem related only to the structure of the graph). In this way, the generality of learned representation can not be guaranteed. 
Recently, contrastive learning has shown attractive potential in unsupervised representation learning, including natural language processing~\cite{radford2019language} and computer vision~\cite{chen2020simple,he2020momentum,grill2020bootstrap,chen2021exploring}. In the graph domain, some works~\cite{you2020graph,velivckovic2018deep,ren2019heterogeneous,suresh2021adversarial,li2022let,ren2021label} also have explored the mechanism, which tends to maximize feature consistency under differently augmented views to learn the desired invariance from these transformations. As a result, data augmentation strategies have a strong influence on contrastive learning performance.
\begin{figure}[t]
	\centering
		\includegraphics[width=\columnwidth]{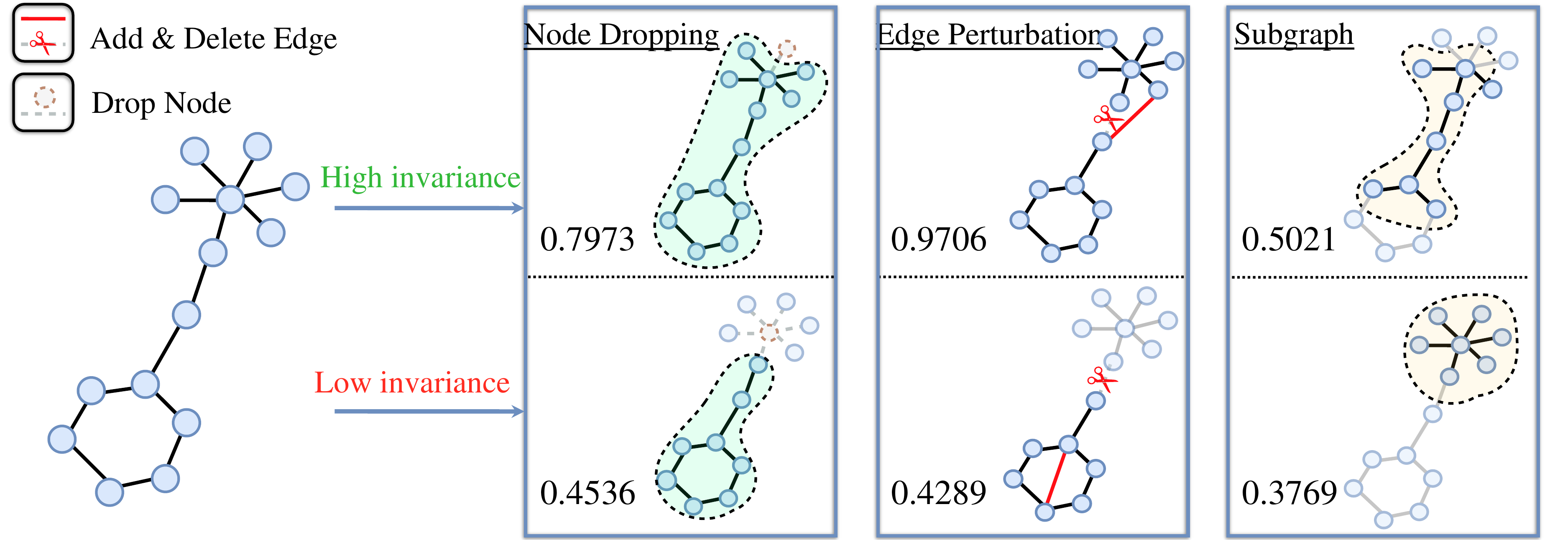}
	\caption{An illustration of the unstable invariance of three graph augmentation strategies. The value of each augmented graph is its similarity to the original graph. The upper part of each augmentation strategy shows augmented graphs preserving high invariance, while the lower part's augmentations bring low invariance.
	}\label{fig:illustration}
\end{figure}

% data aug的劣势（从image和graph差异引入）
Data augmentation has achieved great success in image data where the invariance of various views (e.g., color-invariant, rotation-invariant, and resizing-invariant) are well-understood~\cite{chen2020simple,xiao2020should}. However, due to the complex structural information and the coupling between nodes in the graph, the changes induced by the data augmentation to graphs are not easy to measure. For example, in image data, it is natural to evaluate what kind of data augmentation is more powerful (e.g., color distortion strength for color permuting~\cite{chen2020simple}), but the situation of the graph data is much more complicated. Modifying the attributes of a node is not only related to the target node but also affects its neighbouring nodes. Furthermore, the importance of each node and edge in the graph is far from equivalent, which differs from the importance of pixels in image data. 
Specifically, removing a critical edge is enough to change a graph from a connected graph to a disconnected one, making the augmented graph and the original graph have little learnable invariance. 
In Figure~\ref{fig:illustration}, we provide an illustration to show the unstable invariance between the original graph and augmented graphs under three different data augmentation strategies.
%(Node dropping, Edge perturbation, and Subgraph sampling) 
To measure the destruction of invariance brought by those strategies quantitatively, we annotate the cosine similarity between embeddings of original graph and augmented ones. The graph embeddings are extracted by Node2vec~\cite{grover2016node2vec}.
The upper part of each augmentation strategy shows an augmented graph preserving high invariance, while the lower part's augmentations bring low invariance. Figure~\ref{fig:illustration} shows that the impact of handcrafted augmentations on invariance is uncontrollable, even under the same augmentation strategy. Therefore, commonly used graph augmentation strategies (e.g., Node dropping, Edge perturbation, and Subgraph sampling, etc.) are still not well-explored and lack generalization across datasets from diverse fields. 
% GraphCL
You et al.~\cite{you2020graph} reach similar conclusions after testing various augmentation strategies. For instance, edge perturbation is more suitable for social networks but hurts biochemical molecules. 
% In GraphCL, the composition of data augmentation strategies is an empirical process of finding the invariance suitable for the specific domains, and only a pair-wise composition of strategies can be achieved.
% 很多研究已经在注意到aug选择的重要性和不用aug的可能性
Facing the aforementioned issue, many researchers turn to explore the possibility of discarding data augmentation from contrastive framework recently \cite{lee2022augmentation,yu2022graph,shen2023neighbor}.

To remedy the issue of unstable invariance from inappropriate data augmentations, we propose a novel graph-level contrastive learning framework named CGCL, where no handcrafted graph augmentation is needed. CGCL uses multiple GNN-based graph encoders to enforce contrastive learning in a collaborative way, remedying the limitation of unstable invariance between the original graph and augmented graphs. 
Recalling that contrastive learning aims at learning the invariance across multiple contrastive views, existing GCL methods with data augmentations generate multiple contrastive views from the data augmentations. By contrast, 
%  as shown in Figure 2, 
CGCL generates contrastive views from the encoder perspective. 
Because no handcrafted structural disturbance is injected, contrastive learning on the collaboration of multiple graph encoders can ensure invariance. 
To cope with the problem of model collapse, we devise the asymmetric structure for CGCL. The asymmetry lies in the differences of GNN-based encoders' message-passing schemes. Besides, graph encoders in CGCL are supposed to be complementary for a stronger fitting ability. Specifically, high complementarity indicate that encoders together carry less redundant parameters. For a further theoretical analysis, we propose two metrics: Asymmetry Coefficient~(AC) and Complementarity Coefficient~(CC). Those two metrics are to measure the asymmetry and complementarity of the collaborative framework quantitatively. We validate the performance of CGCL on graph classification task over 9 datasets. Compared with the state-of-the-art methods, CGCL demonstrates better generalization on various datasets and achieves better results without using extra handcrafted data augmentations. In addition, we implement experiments with the two quantitative metrics. The experiments show that the assembly with high asymmetry and complementarity has a better performance, which is in accord with our motivation of designing CGCL's asymmetric architecture and complementary encoders.
The contributions of our work are summarized as follows:
\begin{itemize}
\item We propose a novel \textbf{C}ollaborative \textbf{G}raph \textbf{C}ontrastive \textbf{L}earning (CGCL) to reinforce unsupervised graph-level representation learning, which requires no handcrafted data augmentations. We explain the essence of collaborative framework as generating multiple contrastive views from the encoder perspective.
% \item CGCL shows better generalization across different domains of datasets. The collaboration of multiple graph encoders empowers CGCL with the benefits such as ensemble learning, which enable to learn better representations.
% \item To remedy the issue of unstable invariance from inappropriate data augmentations, CGCL utilizes multiple graph encoders to generate the representations in various embedding spaces as contrastive views, where the original topological and attributes information is preserved, thus guaranteeing the invariance for the graph.
\item We propose the concepts of asymmetric structure and complementary encoders as foundational principles for the collaborative learning paradigm. To provide a more comprehensive theoretical analysis, we put forth two quantitative metrics to assess both the asymmetry and complementarity inherent in the collaborative framework.
\item Extensive experiments on nine datasets show that CGCL has advantages in graph classification task compared with the state-of-the-art methods. Besides, empirical evidence validates that the architecture of CGCL, characterized by its inherent asymmetry and complementarity, indeed yields enhanced performance outcomes. 
\end{itemize}

% \begin{figure}[t]
% 	\centering
%     \includegraphics[width=\columnwidth]{figs/comparison.pdf}
% 	\caption{Comparison of existing GCL methods with data augmentations and the proposed collaborative graph contrastive learning. Graph augmentations in existing methods may bring clumsy and costly pipelines.
% 	}\label{fig:comparison}
% \end{figure}

\section{Related Work} \label{sec:related_work}

\subsection{Graph-level Representation Learning}
Graph-level representation learning is a critical topic~\cite{chen2020graph}, aiming at learning a low-dimensional vector for an entire graph by utilizing its topology structure and nodes feature. 
Early researches could track back to kernel-based methods, including
GK~\cite{shervashidze2009efficient}, 
WL~\cite{shervashidze2011weisfeiler}, and 
DGK~\cite{yanardag2015deep}. Kernel-based methods learn representations with a kernel function measuring the similarity between graph structures. 
Later, the idea of neural networks enables researchers to develop some graph embedding-based methods. Node2vec~\cite{grover2016node2vec} use random walk and skip-gram techniques to capture structural information.
Graph2vec~\cite{narayanan2017graph2vec} learns the graph-level embedding directly.
Sub2vec~\cite{adhikari2018sub2vec} utilizes subgraphs for capturing more global information.
In the recent past, GNNs have shown awe-inspiring capabilities.
%, which aggregate the neighbors' information to learn the latent representations. 
GCN~\cite{kipf2016semi} extends convolution to graphs by a novel Fourier transformation. 
GAT~\cite{velivckovic2017graph} first imports the attention mechanism into graphs.
GIN~\cite{xu2018powerful} develops a general framework to analyze GNN's learning ability. It is worth noting that a pooling function is needed for graph-level representation learning with GNNs~\cite{ju2023comprehensive}. 
Recently, due to the insufficient labeled data, unsupervised setting attracts researchers' attention. Some representative methods include InfoGraph~\cite{sun2019infograph},  
GraphCL~\cite{you2020graph}, 
AD-GCL~\cite{suresh2021adversarial} and 
RGCL~\cite{li2022let}. We will introduce those GCL methods more detailed in the next subsection.

\subsection{Contrastive Learning}
Contrastive learning has been used for unsupervised learning in natural language processing~\cite{radford2019language} and computer vision~\cite{chen2020simple,he2020momentum,grill2020bootstrap,chen2021exploring}.
Inspired by those success, researchers try to generalize contrastive learning to graph data. 
DGI~\cite{velivckovic2018deep} introduces the idea of mutual information maximization~\cite{hjelm2018learning} to learn node-level representation.
InfoGraph~\cite{sun2019infograph} further extends the mutual information maximization to graph-level representations.
% GCC~\cite{qiu2020gcc} utilizes contrastive learning to pre-train a model that can serve for the downstream graph classification task by fine-tuning. 
More recently, GCL methods with data augmentations have achieved the state-of-the-art performance.
GraphCL~\cite{you2020graph} firstly introduces data augmentation strategies for GCL.  
AD-GCL~\cite{suresh2021adversarial} applies an adversarial graph augmentation strategy to avoid capturing redundant information. 
Noticing the importance of augmentation views, RGCL~\cite{li2022let} creates rationale-aware views for graph-level contrastive learning. 
As can be seen, existing graph data augmentation methods heavily rely on the choices of handcrafted augmentations, which may bring unstable performance or clumsy pipelines. 
Some latest researches have been dedicating to explore the the non-necessity of graph data augmentation. 
AF-GRL~\cite{lee2022augmentation} develops an augmentation-free framework, generating views by exploiting nodes sharing the local structural information and global semantics.
SimGCL~\cite{yu2022graph} discards graph augmentations and creates contrastive views by adding noises to embedding space.
NCLA~\cite{shen2023neighbor} uses augmented views automatically learned by graph attention mechanism. 
However, those augmentation-free methods are designed for node-level task, while our proposed CGCL learns graph-level representations through a novel collaborative contrastive framework. Noting that NCLA, which uses multiple differently parameterized GATs to generate contrastive views, is actually a special case of CGCL. Since we consider using GNN-based graph encoders with different schemes rather than the same type of GNNs with different parameters.

% 与上述三个aug-free方法我们的不同之处，贡献也得加上
% 1.graph-level
% 2.我们考虑不同encoder之间的区别，不像NCLA都用不同参数化的GAT

\section{Methodology}\label{sec:method}
In this section, we elaborate on the proposed CGCL framework.
CGCL employs multiple graph encoders to embed the input graphs into various contrastive views.  These graph encoders are dynamically updated in a collaborative manner, predicated upon their respective contrastive losses.

\begin{figure}[t]
		\centering
	\includegraphics[width=\columnwidth]{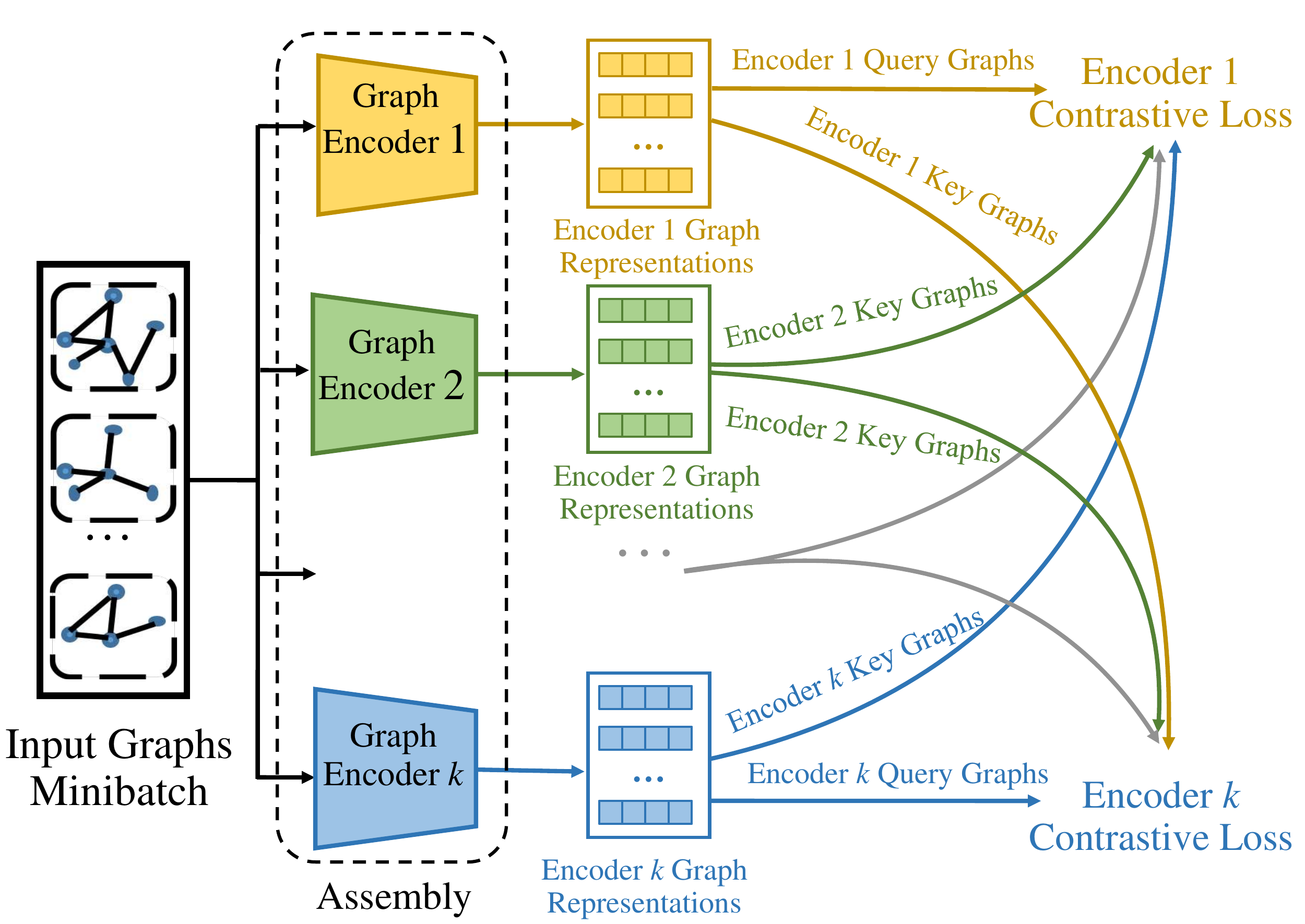}
	\caption{Framework overview of CGCL. Graph Encoder 1, 2, $\cdots$, $k$ embedded the mini-batch graphs into low-dimensional vectors. To optimize the framework collaboratively, each graph encoder calculates its own contrastive loss with the help of others. 
	}\label{fig:overview}
\end{figure}

\subsection{Framework Overview}\label{subsec:graph_overall_framework} 
In Figure~\ref{fig:overview}, we illustrate the overview of CGCL. Multiple graph encoders process input graphs, yielding embeddings for each graph. Every graph encoder updates its parameters by contrasting its learned embeddings to the outputs from the other graph encoders. Specifically, the graph embeddings learned by Graph Encoder $i$ are utilized by Graph Encoder $1, 2,\dots,{i-1}, {i+1},{k}$ as contrastive views. As shown in Figure~\ref{fig:overview}, for example, Graph Encoder 1 uses embeddings learned by itself as query graphs and embeddings learned by all others as key graphs to compute Encoder 1 contrastive loss. After learning, each trained graph encoder can be utilized to extract graph latent representations from a graph when presented as an input.
%for graph representation learning when taking a graph as input. 

\subsection{GNN-based Graph Encoders}\label{subsec:graph_encoder} 
Given a set of graphs, CGCL needs to encode them into vectorized representations. GNNs~\cite{velivckovic2017graph,kipf2016semi,hamilton2017inductive} have demonstrated their outstanding ability in encoding graphs. In CGCL, we mainly employ GNNs as graph encoders. GNNs follow the recursive neighborhood aggregation and certain message-passing scheme~\cite{xu2018powerful} to encode graphs. The aggregation process at the $l$-th layer of a GNN can be represented as:

\begin{equation}\label{eq:gnn}
\begin{aligned}
& \boldsymbol{a}_{n}^{(l)}=\text{AGGREGATION }^{(l)}\left(\left\{\boldsymbol{h}_{n^{\prime}}^{(l-1)}: n^{\prime} \in \mathcal{N}(n)\right\}\right), \\
& \boldsymbol{h}_{n}^{(l)}=\text{COMBINE}^{(l)}\left(\boldsymbol{h}_{n}^{(l-1)}, \boldsymbol{a}_{n}^{(l)}\right), \\
&
\boldsymbol{h}_{\mathcal{G}}=\operatorname{READOUT}\left(\left\{\boldsymbol{h}_{n}^{(l)}: v_{n} \in \mathcal{V}, l \in L\right\}\right)
\end{aligned}
\end{equation}
Here, $\boldsymbol{h}_{n}^{(l-1)}$ is the representation of node $n$ at the $(l-1)$-th layer and $\boldsymbol{h}_{n}^{(0)}$ is the initial feature. $\mathcal{N}(n)$ is the set of neighboring nodes of node $n$. The difference between each GNN is mainly reflected in the use of unique $\text{AGGREGATION}(\cdot)$ and $\text{COMBINE}(\cdot)$ functions, which define the message-passing schemes. For the graph-level representation learning tasks, GNNs need an extra $\text{READOUT}(\cdot)$ to summarize the representation of $\mathcal{G}$ from node representations.

The graph encoder candidates include GIN~\cite{xu2018powerful}, 
GCN~\cite{kipf2016semi},
GAT~\cite{velivckovic2017graph},
DGCNN~\cite{zhang2018end}, 
and Set2Set~\cite{vinyals2015order} in this paper. Technically, any GNN working on graph-level tasks is viable as a graph encoder candidate. Graph encoders can be combined into what we term the CGCL assembly, and subsequently, they can be trained by the collaborative contrastive learning framework
% It is worth noting that the $\operatorname{READOUT}$ of each model should be unified (e.g., global add pooling) instead of using the default method in the original work.

\subsection{Collaborative Contrastive Learning}\label{subsec:collab} 
%We analyze the GNN models selection in Section~\ref{subsubsec:selection} from experimental results.

% Here, we propose three empirical insights first:

% \begin{itemize}[leftmargin=*]
% \item The difference between the embedding spaces of various GNNs is uneven. The performance of the collaboration is harmed when there are too many or too few disparities.
% \item The $\operatorname{READOUT}$ (pooling) method has a significant influence on the graph-level representation. In the supervised learning, pooling methods are strongly sensitive to the specific task, but the selected pooling method may not be suitable for unsupervised representation learning.
% \item Employing more GNNs as graph encoders contributes to better performance. It is equivalent to learning more essential invariance from multi-view representations.
% \end{itemize}

As described in Section~\ref{subsec:graph_encoder}, the proposed CGCL consists of multiple graph encoders. Naturally, the assembly of different types of graph encoders have a significant impact on the performance of CGCL. Therefore, in this section we first explain the essence of collaborative framework. Subsequently, we delve into the two characteristics inherent to CGCL: an asymmetric structure and the complementary encoders. To augment our theoretical exposition, we introduce two quantitative metrics tailored to assess both the asymmetry and complementarity of the collaborative framework.
%[TODO太散了 扩写这个section说了什么]

\subsubsection{\textbf{Encoder Perspective for Generating Contrastive Views.}}\label{subsubsec:essence}
% As shown in Figure~\ref{fig:comparison}, i
In contrast to existing GCL methods which use the same encoder to observe multiple augmented graphs, CGCL uses multiple encoders to observe the same graph and generate contrastive views. It is pivotal to recognize that the quintessence of contrastive learning is to learn invariance between different contrastive views inconsistent to the original graph.This underscores that CGCL innovatively generates different contrastive views from the encoder perspective, eschewing traditional data augmentations. Given that variations in either encoder configurations or data invariably manifest in the resultant embeddings, collaborative contrastive learning parallels the efficacy of augmentation-based methods but is free from the often cumbersome or resource-intensive augmentation phase. With the multi contrastive views, encoders collaboratively interact. Each encoder's output, serving as both query and key graphs, influences its inherent loss and those of its counterparts. This process fosters a mutual distillation of knowledge on graph latent representations.

% \subsubsection{\textbf{Collaborative Distillation.}}
% As illustrated in Figure~\ref{fig:overview}, CGCL employs different graph encoders in a collaborative way. We refer this process as collaborative distillation since each graph encoder calculates its own contrastive loss with the help of other encoders. Specifically, every encoder's outputs serve both query and key graphs when computing its own loss and others' losses. During the collaborative process, knowledge of graph latent representations learned from different encoders is distilled between each other. 

% \subsubsection{\textbf{Asymmetry and Complementarity.}}\label{subsubsec:asyandcom}
% Now we introduce two properties of the proposed collaborative framework. To be more specific, CGCL have not only an asymmetric architecture but also complementary graph encoders. The detailed analysis is described below. 
% \subsubsection{\textbf{Asymmetric Architecture.}}\label{subsubsec:asy}

\begin{figure}[t]
		\centering
	\includegraphics[width=\columnwidth]{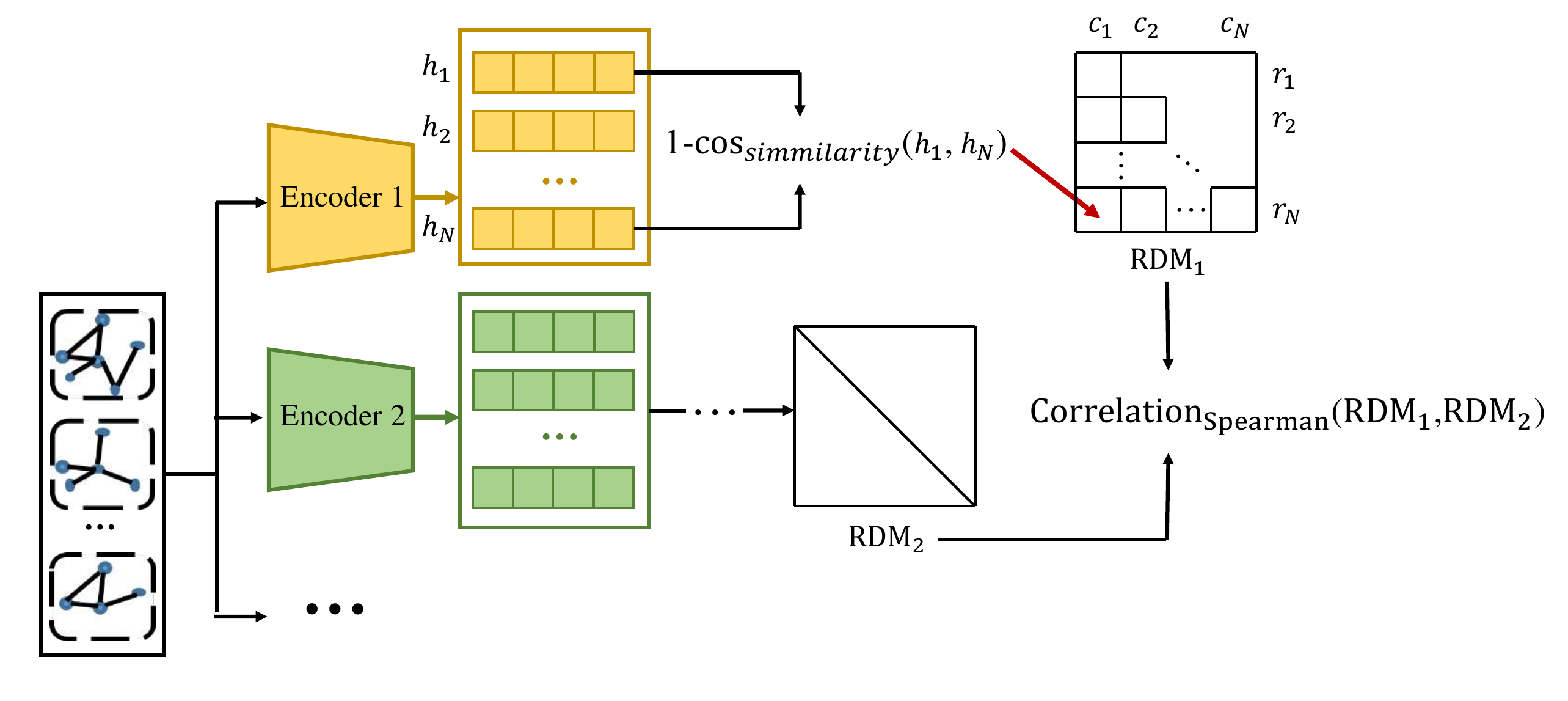}
	\caption{Calculation of correlation between RDMs. With $k$ encoders, Asymmetry Coefficient is calculated by averaging the correlation between any pair of encoders. 
	}\label{fig:rdm}
\end{figure}

\subsubsection{\textbf{Asymmetric Architecture.}} \label{subsubsec:asyandcom}
% For contrastive learning, a critical issue is model collapse. Existing methods~\cite{chen2020simple,he2020momentum,grill2020bootstrap,chen2021exploring} adopt various techniques to alleviate such problem. 
In contrastive learning, model collapse presents a significant challenge. Existing methods, such as BYOL~\cite{grill2020bootstrap} and SimSiam~\cite{chen2021exploring}, employ techniques like the Exponential Moving Average and Stop Gradient respectively to avoid model collapse.
Those techniques can be concluded as introducing asymmetry into model architecture. Reflecting on CGCL, diverse GNN-based graph encoders with distinct message-passing schemes are employed to ensure an asymmetric architecture. Essentially, the variance in these schemes introduces the desired asymmetry. Thus, CGCL's assembly necessitates multiple encoder types. It is noted that NCLA~\cite{shen2023neighbor}, by employing differently parameterized GATs as encoders, essentially adheres to a symmetric architecture, potentially predisposing it to model collapse. Conversely, CGCL's asymmetric design inherently mitigates such risks.
As discussed in the above analysis, diverse encoders sustain the asymmetry within CGCL's assembly. Naturally, the extent of asymmetry can be reflected on the differences between encoders. Thus, we introduce the Representational Dissimilarity Matrix~(RDM), which is a prevalent tool in cognitive neuroscience~\cite{popal2019guide}. Model discrepancies can be measured by calculating the correlation between two RDMs. For clarification, we illustrate the process of calculating the correlation between RDMs in Figure~\ref{fig:rdm}. Firstly, with $k$ trained graph encoders and graph data with batch size $N$, we encode the data into $k$ groups of batch embeddings. Then, for each batch, we calculate the cosine distance between the pairwise embeddings. We can construct an $N$\texttimes$N$ RDM based on the pairwise distances. Finally, we calculate the correlation between those RDMs pairwise. In this work, we use Spearman correlation to avoid the assumption of a linear match between the RDMs.
%为了量化不对称，提出一个指标，介绍下rdm，为什么这个指标能反映表示学习模型差异性，进而反映assembly的不对称，最后提一下反例情况，即如果完全对称，encode都是同一种，理论情况下会收敛到一样情况，该指标会趋于0.
Based on the concept of RDM, we define the Asymmetry Coefficient of a certain CGCL's assembly as follows: 
\begin{definition}
\textit{Let k denote the number of trained encoders, and $\text{RDM}_{i}^{D}$ be the Representational Dissimilarity Matrix defined by encoder $i$ and data $D$. Then the Asymmetry Coefficient~(AC) is formulated as:}
\begin{equation}
AC=1-\frac{1}{C_{k}^{2}} \sum_{(i,j)\in \text{Combination}(k)}^{} \text{Correlation} (\text{RDM}_{i}^{D},\text{RDM}_{j}^{D})
\label{eq:asy}
\end{equation}
\end{definition}
A high AC indicates that CGCL's assembly holds high asymmetry and vice versa. As an illustrative extreme, if GNN-based encoders were homogeneous, their parameters would converge due to identical message-passing schemes and network structures. With congruent trained parameters, $\text{RDM}_{i}^{D}$ and $\text{RDM}{j}^{D}$ would be roughly the same, resulting in a correlation nearing 1. Given this context, the eventual AC would approach 0, underscoring the absence of asymmetry in an assembly comprising a singular encoder type.

% \subsubsection{\textbf{Complementary Encoders.}}\label{subsubsec:com}
\subsubsection{\textbf{Complementary Encoders.}}
% \paragraph{Complementary Encoders.} 
Within CGCL, multiple graph encoders observe input graphs to yield contrastive views. Ideally, these encoders should exhibit complementarity to enhance fitting capability. Specifically, an assembly with encoders possessing non-redundant observation angles demonstrates high complementarity. Redundancies in observation angles can be inferred from overlapping encoder parameters. This notion of complementarity in CGCL mirrors the diversity imperative of base learners in ensemble learning, where distinct learners better capture varied information.
As discussed in Section~\ref{subsubsec:essence}, CGCL generates contrastive views from encoder perspective, distinguishing it from the data augmentations in traditional GCL methods. Inspired by \cite{gontijo2020affinity}, we introduce a loss-centric metric to measure the complementarity of CGCL's encoders. For clarification, we refer the training loss upon completion as the stopping loss. Given a consistent dataset, a smaller stopping loss signifies enhanced assembly fitting ability. This capability is directly proportional to the non-redundant parameters across all encoders. As previously highlighted, complementarity is evident through non-redundant observation angles. With the above intuition, we define the Complementarity Coefficient of a certain CGCL's assembly as follows:
\begin{definition}
\textit{Let k denote the number of encoders, and $L_i^{\text{Stop}}$ be the loss of encoder $i$ when collaborative training is stopped. The Complementarity Coefficient~(CC) is given by:}
\begin{equation}
CC=-\frac{1}{k} \sum_{i\in 1,2,\dots,k}{L_i^{\text{Stop}} }
\label{eq:com}
\end{equation}
\end{definition}
With the above definition, CGCL's encoders are more complementary with a higher CC. In Equation~(\ref{eq:com}), we introduce normalization to get rid of the effect of $k$.

\subsubsection{\textbf{Discussion.}}
% \paragraph{Discussion.}
The assembly of CGCL is designed to exhibit both asymmetry and complementarity. For a more precise quantitative evaluation, we introduce two metrics tailored to assess these properties of the collaborative framework. Importantly, the computation of these metrics is solely based on pre-training stage and remains uninfluenced by downstream classifiers, rendering it a valuable guidance for assembly selection.

%除了不对称，还要考虑互补性，这块可以扣不同视角观察的理论。intuition：同个数据集loss小<-模型拟合能力强<-assembly非冗余参数多<-assembly内部encoders之间互补性好

% \subsection{Graph representations as contrastive views}\label{subsec:augment} 
% The graph representations learned by different GNNs-based encoders essentially act the contrastive views in CGCL. However, CGCL differs from GraphCL~\cite{you2020graph} where contrastive views are generated by applying manually augmentations on graph-structured data. CGCL generates contrastive views through different "message-passing" schemes, equivalent to observing a graph from multiple views. 
% The graph representations learned by different GNNs essentially act the data augmentation in CGCL. However, the data augmentation mechanism used here differs from that of GraphCL~\cite{you2020graph}, which is designed manually based on priors of graph-structured data. CGCL achieves this kind of data augmentation through different "message-passing" schemes, equivalent to observing a graph from multiple views. 
%We provide the detailed analysis of this insight in Section~\ref{subsec:theo} in a theoretical way.

\subsection{Batch-wise Contrastive Loss}\label{subsec:graph_contrastive_loss} 
We
utilize the training batch to conduct contrastive learning. Given a minibatch of $N$ graphs $\{\mathcal{G}_1, \mathcal{G}_2, \dots, \mathcal{G}_N\}$ as the input of $k$ graph encoders $\{\mathit{M}_1, \mathit{M}_2,\dots, \mathit{M}_k\}$. Contrastive learning can be considered as learning an encoder for a dictionary look-up task~\cite{he2020momentum}. We take $\mathit{M}_1$ as an example. The representations $\{\boldsymbol{h}^{\mathit{M}_1}_{\mathcal{G}_1}, \boldsymbol{h}^{\mathit{M}_1}_{\mathcal{G}_2},\dots, \boldsymbol{h}^{\mathit{M}_1}_{\mathcal{G}_N}\}$ learned by $\mathit{M}_1$ act as query graphs, while the representations learned by other encoders are key graphs. Between query and key graphs, the positive graph representation pairs are:

\begin{equation}
\begin{aligned}
\{(\boldsymbol{h}^{\mathit{M}_1}_{\mathcal{G}_i}, \boldsymbol{h}^{\mathit{M}_j}_{\mathcal{G}_i})\ |\  i\in[1,N], j\in[2,k]\}
\end{aligned}
\end{equation}
Therefore, each graph encoded by $\mathit{M}_1$ has $k-1$ positive samples. In essence, the positive graph pair indicates the representations of the same graph but learned by different graph encoders. Meanwhile, the other pairs are negative. Here, we utilize the InfoNCE~\cite{oord2018representation} to calculate the contrastive loss for each graph encoder. We define the contrastive loss between two graph encoders $\mathit{M}_p$ and $\mathit{M}_q$ first, so the contrastive loss $\mathcal{L}_{\mathit{M}_p}(\mathit{M}_q)$ can be represented as:

\begin{equation}
\begin{aligned}
\mathcal{L}_{\mathit{M}_p}(\mathit{M}_q)=-\sum_{j=1}^{N}\log \frac{\exp \left(\boldsymbol{h}^{\mathit{M}_p}_{\mathcal{G}_j} \cdot \boldsymbol{h}^{\mathit{M}_q}_{\mathcal{G}_j} / \tau\right)}{\sum_{n=1}^{N} \exp \left(\boldsymbol{h}^{\mathit{M}_p}_{\mathcal{G}_j} \cdot \boldsymbol{h}^{\mathit{M}_q}_{\mathcal{G}_n} / \tau\right)}
\end{aligned}
\end{equation}
Here, $\tau$ is the temperature hyper-parameter.
In this contrastive loss, $\mathit{M}_p$ acts as the query encoder, while $\mathit{M}_q$ provides key graphs for contrast. When considering all collaborative graph encoders, the contrastive loss of $\mathit{M}_p$ can be denoted as:

\begin{equation}\label{eq:loss}
\begin{aligned}
\mathcal{L}_{\mathit{M}_p} = \displaystyle\sum_{\mathit{M}_q \in \{\mathit{M}_i\}_{i=1}^k \setminus {\mathit{M}_p}} \mathcal{L}_{\mathit{M}_p}( \mathit{M}_q)
\end{aligned}
\end{equation}
$\mathit{M}_p$ can be updated based on the contrastive loss $\mathcal{L}_{\mathit{M}_p}$ through back-propagation. Other graph encoders follow the same process to compute their own contrastive loss and get updated by minibatch training iteratively. The detailed collaborative learning process of CGCL is in Algorithm~\ref{alg:collaborative_learning}.

\begin{algorithm}[tb]
	\caption{Collaborative Graph Contrastive Learning} 
	%	\hspace*{0.02in} {\bf Input:} %算法的输入， \hspace*{0.02in}用来控制位置，同时利用 \\ 进行换行
	%	\hspace*{0.02in} {\bf Output:} 
	\begin{algorithmic}[1] 
		\REQUIRE
		The set of GNN-based graph encoders $\{\mathit{M}_i\}_{i=1}^k$; Batch size $N$; Temperature parameter $\tau$
		\ENSURE
		Trained graph neural networks $\{\mathit{M}_i\}_{i=1}^k$;

		\FORALL {sampled graph minibatch $\{\mathcal{G}_j\}_{j=1}^N$}
		\FOR {$j \in \{1,\dots,N\}$}
		\FORALL {$\mathit{M}_i \in \{\mathit{M}_i\}_{i=1}^k$}
		\STATE $\boldsymbol{h}^{\mathit{M}_i}_{\mathcal{G}_j} = \mathit{M}_i(\mathcal{G}_j)$  %$\ \ \ \ \ \ $//Encoding the input graph
		\ENDFOR
		\ENDFOR
		\FORALL {$\mathit{M}_p \in \{\mathit{M}_i\}_{i=1}^k$}
		\FORALL {$\mathit{M}_q \in \{\mathit{M}_i\}_{i=1}^k \setminus {\mathit{M}_p}$}	
		%\STATE define contrastive loss between two encoders:
		\STATE $\mathcal{L}_{\mathit{M}_p}( \mathit{M}_q) = -\sum_{j=1}^{N}\log \frac{\exp \left(\boldsymbol{h}^{\mathit{M}_p}_{\mathcal{G}_j} \cdot \boldsymbol{h}^{\mathit{M}_q}_{\mathcal{G}_j} / \tau\right)}{\sum_{n=1}^{N} \exp \left(\boldsymbol{h}^{\mathit{M}_p}_{\mathcal{G}_j} \cdot \boldsymbol{h}^{\mathit{M}_q}_{\mathcal{G}_n} / \tau\right)}$
		 \ENDFOR	
		
		\STATE $\mathcal{L}_{\mathit{M}_p} = \displaystyle\sum_{\mathit{M}_q \in \{\mathit{M}_i\}_{i=1}^k \setminus {\mathit{M}_p}} \mathcal{L}_{\mathit{M}_p}( \mathit{M}_q)$
		\STATE update graph encoder $\mathit{M}_p$ to minimize $\mathcal{L}_{\mathit{M}_p}$ %\COMMENT{parallel}
		\ENDFOR
		\ENDFOR
		\STATE \RETURN graph neural networks $\{\mathit{M}_i\}_{i=1}^k$;
	\end{algorithmic}\label{alg:collaborative_learning}
\end{algorithm}

\section{Experiments}\label{sec:experiment}

In this section, we first introduce the datasets and baselines. To verify the effectiveness of proposed CGCL, we conduct experiments to investigate following research questions:
\begin{itemize}[leftmargin=*]
\item\textbf{RQ1:} How does CGCL perform on unsupervised graph representation learning as compared to other state-of-art methods?
% \item\textbf{RQ2:} How does CGCL perform under the setting of pretrain-finetune learning?
\item\textbf{RQ2:} How do the assembly of graph encoders impact the performance of CGCL concerning the aspects of asymmetry and complementarity 
\item\textbf{RQ3:} How do the multiple graph encoders in CGCL converge in the collaborative learning process?
% \item\textbf{RQ5:} What is the time cost of CGCL as compared to other baseline GCL methods?
% \item\textbf{RQ5:} What is the visualization of embeddings learned by CGCL?
\end{itemize}
%graph-level classification in the setting of unsupervised learning and pretrain-finetune learning~\cite{you2020graph}. In addition, we analyze the assembly of graph encoders in CGCL and visualize the learned graph representations. At last, we verify the reliability of the collaboration between graph encoders through convergence analysis.
%Then we demonstrate the node representations learned by CGCL in the task of node classification~\cite{velivckovic2018deep}. 
\subsection{Datasets}
To evaluate CGCL for graph-level representation learning, we conduct experiments on 9 widely used benchmark datasets: NCI1, PROTEINS, D$\&$D,	MUTAG, COLLAB, IMDB-BINARY, IMDB-MULTI, REDDIT-BINARY, REDDIT-MULTI-5K. All of them come from the benchmark TUDataset~\cite{morris2020tudataset}. 
%For node-level representation learning, we test CGCL on 3 well-known benchmark dataset: Cora, CiteSeer and Pubmed~\cite{sen2008collective}. For all the above datasets, we reach them with the support of the PyTorch Geometric Library. 

\subsection{Baselines}
Following~\cite{you2020graph}, we select three categories of state-of-the-art methods as our baselines: 
3 kernel-based methods including 
Weisfeiler-Lehman subtree kernel (\textbf{WL})~\cite{shervashidze2011weisfeiler}, 
Graphlet Kernel (\textbf{GK}), 
Deep Graph Kernel (\textbf{DGK})~\cite{yanardag2015deep}, 
3 graph embedding-based methods including \textbf{Sub2vec}~\cite{adhikari2018sub2vec},
\textbf{Node2vec}~\cite{grover2016node2vec},
\textbf{Graph2vec}~\cite{narayanan2017graph2vec} and 4 GCL methods including  \textbf{InfoGraph}~\cite{sun2019infograph}, 
\textbf{GarphCL}~\cite{you2020graph},
\textbf{AD-GCL}~\cite{suresh2021adversarial}, 
\textbf{RGCL}~\cite{li2022let}.
 %They extend document embedding neural networks to learn representations of entire graphs. 
%They first decompose graphs into sub-components based on the kernel definition, then learn graph embeddings in a feature-based manner.
%It is an unsupervised with the pooling operator for graph representation learning base on the mutual information maximization.
	
 %: It follows pre-training and fine-tuning paradigm for graph representation learning. 
 %: It implements contrastive learning between augmented graphs to obtain graph representation in an unsupervised manner.

%-----------------------------------------------------------------------
%\subsection{Graph-level Representation Learning}
%\subsubsection{Unsupervised Graph Representation Learning}

\subsection{Unsupervised Graph Representation Learning (RQ1)}

\begin{table*}[t]
	%\footnotesize
        %\scriptsize
	\centering
        \caption{Performance on graph classification accuracy in the setting of unsupervised learning. The best result is \textbf{bolded}.} % and the runner-up is \underline{underlined}
	\begin{tabular}{l| c| c| c| c| c}
	\hline           \textbf{Methods}&NCI1&PROTEINS&D$\&$D&MUTAG&COLLAB\\
		%			\cline{2-6}
		%			&Accuracy&F1 Score&Accuracy&F1 Score&Accuracy\\
		\hline
            WL&76.65$\pm$1.99&72.92$\pm$0.56&76.44$\pm$2.35&80.72$\pm$3.00& - \\
		GK&62.48$\pm$2.11& 72.23$\pm$4.49 & 72.54$\pm$3.83 & 81.66$\pm$2.11 & 72.84$\pm$0.28\\
		DGK&	\textbf{80.31$\pm$0.46} & 73.30$\pm$0.82 & 71.12$\pm$0.21 & 87.44$\pm$2.72 & 73.09$\pm$0.25\\
		\hline
		Node2vec&54.89$\pm$1.61 & 57.49$\pm$3.57 & 67.12$\pm$4.32 & 72.63$\pm$10.20 & - \\
		Sub2vec&52.84$\pm$1.47 & 53.03$\pm$5.55 & 59.34$\pm$8.01 & 61.05$\pm$15.80 & - \\
		Graph2vec&73.22$\pm$1.81 & 73.30$\pm$2.05 & 71.98$\pm$3.54 & 83.15$\pm$9.25 & -  \\
		\hline
		InfoGraph&76.20$\pm$1.06 & 74.44$\pm$0.31 & 72.85$\pm$1.78 & 89.01$\pm$1.13 & 70.65$\pm$1.13 \\
		%\hline
		GraphCL&77.87$\pm$0.41 & 74.39$\pm$0.45 & 78.62$\pm$0.40 & 86.80$\pm$1.34 & 71.36$\pm$1.15\\
            AD-GCL&73.91$\pm$0.77 & 73.28$\pm$0.46&75.79$\pm$0.87& 88.74$\pm$1.85&72.02$\pm$0.56\\
            RGCL& 78.14$\pm$1.08&75.03$\pm$0.43 &78.86$\pm$0.48 & 87.66$\pm$1.01& 70.92$\pm$0.65\\
		%			\hline
		\hline
		${\text{CGCL}}_{GIN}$&77.89$\pm$0.54 & \textbf{76.28$\pm$0.31} & \textbf{79.37$\pm$0.47} & \textbf{89.05$\pm$1.42} & \textbf{73.28$\pm$2.12}\\

	\hline	
        \end{tabular} 
    \begin{tabular}{l| c| c| c| c}
	\hline           \textbf{Methods}&IMDB-B&IMDB-M&RDT-B&RDT-M\\
		%			\cline{2-6}
		%			&Accuracy&F1 Score&Accuracy&F1 Score&Accuracy\\
		\hline
            WL&72.30$\pm$3.44&46.95$\pm$0.46&68.82$\pm$0.41&46.06$\pm$0.21\\
		GK& 65.87$\pm$0.98 & - & 77.34$\pm$0.18 & 41.01$\pm$0.17\\
		DGK& 66.96$\pm$0.56 & 44.55$\pm$0.52 & 78.04$\pm$0.39 & 41.27$\pm$ 0.18\\
		\hline
		Node2vec& 61.03$\pm$7.13 & - & - & - \\
		Sub2vec& 55.26$\pm$1.54 & 36.67$\pm$0.83 & 71.48$\pm$0.41 & 36.68$\pm$0.42 \\
		Graph2vec& 71.10$\pm$0.54 & 50.44$\pm$0.87 & 75.78$\pm$1.03 & 47.86$\pm$0.26 \\
		\hline
		InfoGraph &73.03$\pm$0.87 & 49.69$\pm$0.53 & 82.50$\pm$1.42 & 53.46$\pm$1.03 \\
		%\hline
		GraphCL&71.14$\pm$0.44 & -  & 89.53$\pm$0.84 & 55.99$\pm$0.28 \\
            AD-GCL&70.21$\pm$0.68& 46.60$\pm$0.01 &90.07$\pm$0.85 & 54.33$\pm$0.32 \\
            RGCL&71.85$\pm$0.84  & 49.27$\pm$0.00  & 90.34$\pm$0.58&\textbf{56.38$\pm$0.40}  \\
		%			\hline
		\hline
		${\text{CGCL}}_{GIN}$ & \textbf{73.11$\pm$0.74} & \textbf{51.73$\pm$1.37} & \textbf{91.31$\pm$1.22} & 54.47$\pm$1.08 \\

	\hline	
	\end{tabular} 
	\label{tab:results_unsupervised}
\end{table*}

We first evaluate the representations learned by CGCL in the setting of unsupervised learning. We follow the same process of InfoGraph~\cite{sun2019infograph}, where representations are learned by models without any labels and then fed into a SVM to evaluate the graph classification performance. 
${\text{CGCL}}_{GIN}$ denotes GIN~\cite{xu2018powerful} trained under the proposed framework CGCL. The graph encoder candidates include  GAT, GCN, and DGCNN, which are combined with GIN to conduct collaborative contrastive learning. The representations learned by ${\text{CGCL}}_{GIN}$ are extracted for the downstream graph classification. We test the task 10 times and report the average accuracy and the standard deviation. In our experiments, only two graph encoders or three graph encoders collaborative learning settings are used. The result of ${\text{CGCL}}_{GIN}$ we report in Table~\ref{tab:results_unsupervised} is the best. From the results in Table~\ref{tab:results_unsupervised}, CGCL achieves the best result on seven out of nine datasets. Noting that GraphCL and RGCL have better result on RDT-M, where average node degree is extremely large compared to other datasets. On the contrary, those two methods with data augmentations perform worse then the proposed CGCL on other datasets with a low average node degree. This observation is in line with our illustration in Figure~\ref{fig:illustration}, since when the graph is small and not dense enough, the perturbation from the data augmentation is more likely to damage the invariance, e.g., a greater probability of affecting the graph connectivity by dropping a single node.

%\subsubsection{Pretrain-Finetune Graph Representation Learning}
% \subsection{Pretrain-Finetune Graph Representation Learning (RQ2)}

% In this setting, CGCL first conducts collaborative contrastive learning to pretrain the graph encoders. Then we use 10\% labeled graph data of the whole dataset to finetune the pretrained graph encoder end-to-end for the downstream graph classification tasks. Here, we utilize GAE~\cite{kipf2016variational}, InfoGraph~\cite{sun2019infograph}, GCC~\cite{qiu2020gcc}, and GraphCL~\cite{you2020graph} to pretrain GCN for comparison. GAE focuses on the graph structure reconstruction and InfoGraph utilizes the local-glocal mutual information maximization. GCC and GraphCL pretrain the graph encoder based on the contrastive learning between augmented graph data. CGCL employs GCN and GIN as graph encoders to fulfill the designed contrastive learning in each dataset. For fairness, we finetune GCN with 10\% labeled graph and report the graph classification accuracy. More implementation details can be referred to in Appendix. The performance in each dataset of different methods is presented in Table~\ref{tab:results_semi}. As can be seen, CGCL has achieved the best results on 4 datasets, and the proposed method saves time cost in generating contrastive views and exploring to composite those views in an appropriate strategy.

\subsection{Asymmetry and Complementarity of CGCL's Assembly~(RQ2)}\label{subsubsec:selection}

\begin{figure*}[t]	    
	\centering
    \includegraphics[width=\linewidth]{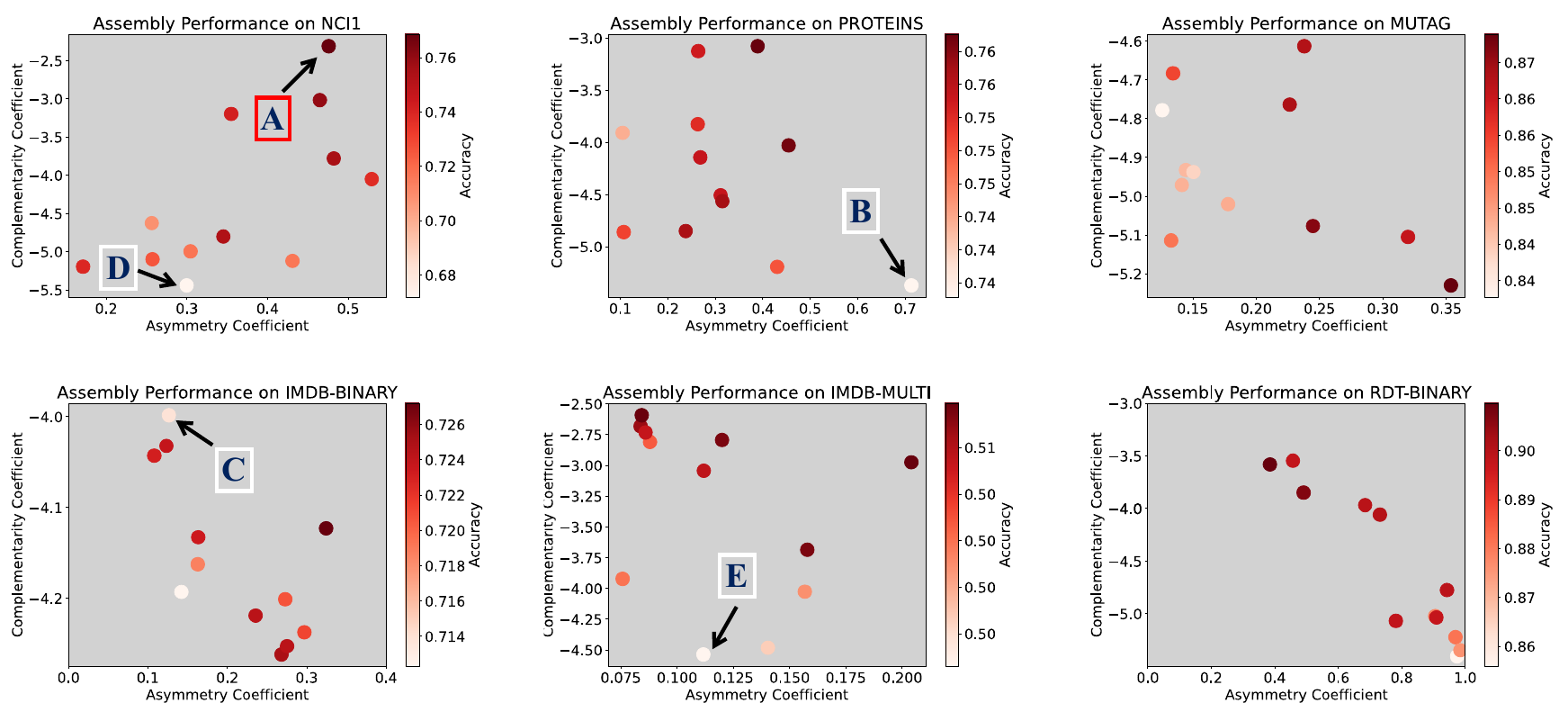}%
	\caption{The performance of CGCL's assembly with respect to asymmetry and complementarity over multiple datasets. Point A indicates an example of the assembly with a high AC and CC. While point B has a high AC and a low CC, point C is with a low AC and a high CC high. Point D an E refer to the examples whose AC and CC are both low.}\label{fig:assembly}
\end{figure*}

For a further exploration of CGCL's working mechanism, we test various assembly with respect to the two quantitative metrics AC and CC. The candidate encoders include GIN, GCN, GAT and Set2Set. Specifically, we combine two or three of graph encoders on six datasets and use the best result of multiple encoders as this assembly's result. Each point represents a certain assembly and its color indicates the performance. This experiment is implemented in the setting of unsupervised graph representation learning. 
As illustrated in Figure~\ref{fig:assembly}, the best assembly generally appears in the top right-hand corner, for example point A. This indicates that the assembly with both high AC and CC generally performs better, which further justifies the rationality of the design of asymmetric architecture and complementary encoders for CGCL. Besides, the assembly holding only one property may fail the graph representation learning and result in the bad performance of downstream classification.
For instance, the assembly in the bottom right-hand corner~(point B) on PROTEINS has high asymmetry, however performs bad because of its very low complementarity.
Another bad example is the assembly in the top left-hand corner of IMDB-BINARY~(point C). Though it has a high CC, the AC of this assembly is fairly low. In addition, there also exist assembly, whose AC and CC are both at low levels, such as point D on NCI1 and point E on IMDB-MULTI. With the above analysis, we further confirm the effectiveness of our proposed collaborative framework.

\subsection{Convergence Analysis (RQ3)}

\begin{figure}[t]	    
		\centering
		\includegraphics[width=\columnwidth]{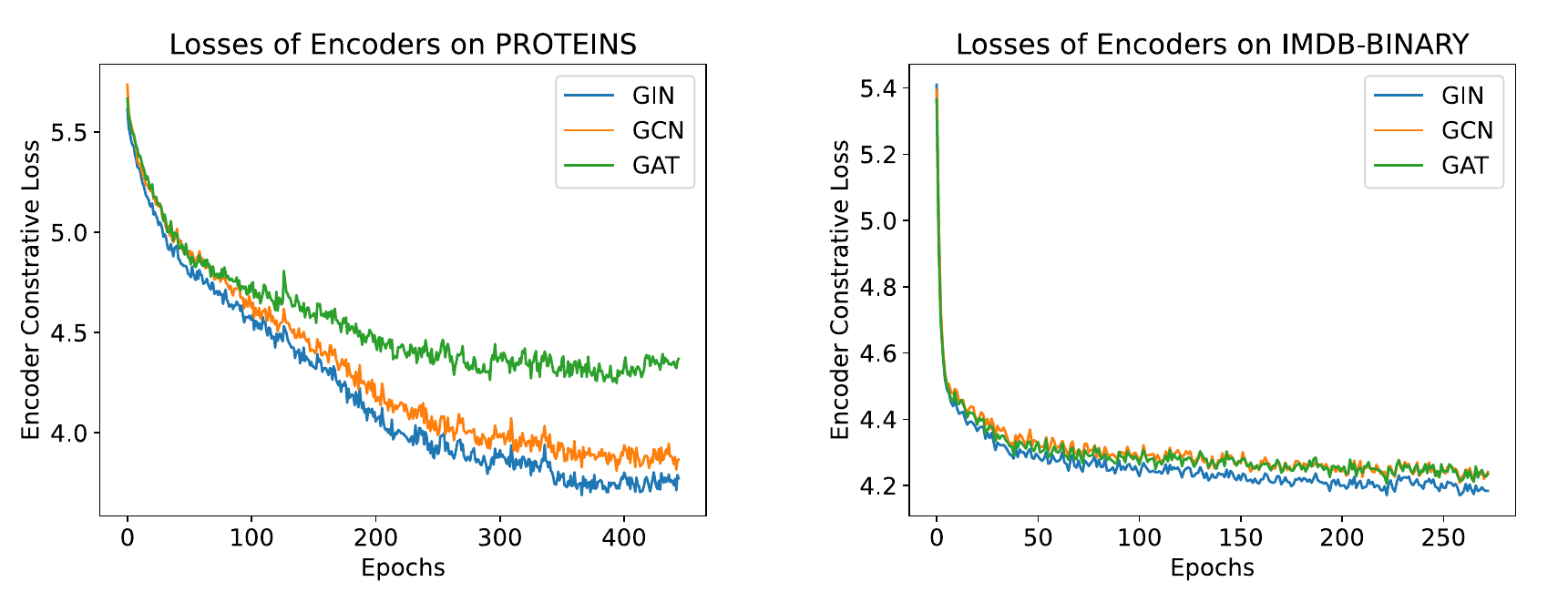}%
	\caption{Empirical convergence study of different graph encoders in $\text{CGCL}$ on PROTEINS and IMDB-BINARY. }\label{fig:loss}
\end{figure}

In CGCL, multiple graph encoders compute their own contrastive losses based on representations learned by others, and optimize their losses collaboratively. To check the reliability of collaborative mechanism, we empirically analyze the convergence in the optimization process of each individual encoder on PROTEINS and IMDB-BINARY. The assembly we use includes GIN, GCN and GAT. In Figure~\ref{fig:loss}, we notice that each graph encoder converges synchronously on the two datasets, which justifies our proposed collaborative learning framework. For a further analysis, we list the RDMs correlation between pairs of GIN, GCN and GAT in Table~\ref{tab:loss} for reference. According to the figure and the table, the trend of three encoders' contrastive losses is in accord with their RDMs' correlation of each other. In terms of PROTEINS, for example, the correlation between GIN and GCN is relatively high~(0.8620), while the correlations between GAT and those two are low and close~(0.5732/0.6308). We observe that those relations correspond well with the change of losses, which also exhibits the distinction of GAT. Such accordance further proves the rationality of Asymmetry Coefficient proposed in Section~\ref{subsubsec:asyandcom}.

\begin{table}[t]
  \centering
  \caption{The RDMs Correlation between graph encoders in ${\text{CGCL}}_{GIN/GCN/GAT}$ on PROTEINS and IMDB-BINARY.}
    \begin{tabular}{l|c|c}
    \hline
    \textbf{RDMs Correlation} & PROTEINS & IMDB-B \\
    \hline
    GIN-GCN & 0.8620 & 0.7164 \\
    GIN-GAT & 0.5732 & 0.6891 \\
    GCN-GAT & 0.6308 & 0.7038 \\
    \hline
    \end{tabular}%
  \label{tab:loss}%
\end{table}%

% \subsection{Time Cost Comparison (RQ5)}

% \subsection{Visualization of Learned Representations (RQ5)}\label{sec:visual}
% In Figure~\ref{fig:TSNE}, we visualize the graph-level embeddings learned by ${\text{CGCL}}_{GIN}$ in the unsupervised setting on PROTEINS dataset. To intuitively see how CGCL works, Node2vec~\cite{grover2016node2vec} is used for a comparison. We run t-SNE~\cite{buja1996interactive} and annotate each graph by its class label. It's obvious that ${\text{CGCL}}_{GIN}$ have learned better embeddings than Node2vec, though both of them are unsupervised methods. The visualization further shows our proposed CGCL's effectiveness.

\section{Conclusion}\label{sec:conclusion}

In this study, we introduce CGCL, a novel collaborative graph contrastive learning framework, designed to address the invariance challenge encountered in current GCL methods. Unlike the conventional practice of constructing augmented graphs by hand, CGCL employs multiple GNN-based encoders to generate multiple contrastive views. This obviates the need for explicit structural augmentation perturbations, thus ensuring invariance. Graph encoders of CGCL learn the graph representations collaboratively, and enhance each other's learning ability in an unsupervised manner. We then propose the concepts of asymmetric structure and complementary encoders as the design principle for the collaborative framework. For a further theoretical analysis, we propose two quantitative metrics to measure the asymmetry and complementarity of the collaborative framework.
Extensive experiments substantiate the advantages of CGCL and underscores the potential of collaborative framework in the field of GCL.

\bibliographystyle{splncs04}
\bibliography{ref}

\begin{thebibliography}{10}
\providecommand{\url}[1]{\texttt{#1}}
\providecommand{\urlprefix}{URL }
\providecommand{\doi}[1]{https://doi.org/#1}

\bibitem{adhikari2018sub2vec}
Adhikari, B., Zhang, Y., Ramakrishnan, N., Prakash, B.A.: Sub2vec: Feature
  learning for subgraphs. In: Pacific-Asia Conference on Knowledge Discovery
  and Data Mining. pp. 170--182. Springer (2018)

\bibitem{chen2020graph}
Chen, F., Wang, Y.C., Wang, B., Kuo, C.C.J.: Graph representation learning: a
  survey. APSIPA Transactions on Signal and Information Processing  \textbf{9}
  (2020)

\bibitem{chen2020simple}
Chen, T., Kornblith, S., Norouzi, M., Hinton, G.: A simple framework for
  contrastive learning of visual representations. In: International conference
  on machine learning. pp. 1597--1607. PMLR (2020)

\bibitem{chen2021exploring}
Chen, X., He, K.: Exploring simple siamese representation learning. In:
  Proceedings of the IEEE/CVF conference on computer vision and pattern
  recognition. pp. 15750--15758 (2021)

\bibitem{gontijo2020affinity}
Gontijo-Lopes, R., Smullin, S.J., Cubuk, E.D., Dyer, E.: Affinity and
  diversity: Quantifying mechanisms of data augmentation. arXiv preprint
  arXiv:2002.08973  (2020)

\bibitem{grill2020bootstrap}
Grill, J.B., Strub, F., Altch{\'e}, F., et~al.: Bootstrap your own latent-a new
  approach to self-supervised learning. Advances in neural information
  processing systems  \textbf{33},  21271--21284 (2020)

\bibitem{grover2016node2vec}
Grover, A., Leskovec, J.: node2vec: Scalable feature learning for networks. In:
  Proceedings of the 22nd ACM SIGKDD international conference on Knowledge
  discovery and data mining. pp. 855--864 (2016)

\bibitem{hamilton2017inductive}
Hamilton, W.L., Ying, R., Leskovec, J.: Inductive representation learning on
  large graphs. arXiv preprint arXiv:1706.02216  (2017)

\bibitem{he2020momentum}
He, K., Fan, H., Wu, Y., Xie, S., Girshick, R.: Momentum contrast for
  unsupervised visual representation learning. In: Proceedings of the IEEE/CVF
  Conference on Computer Vision and Pattern Recognition. pp. 9729--9738 (2020)

\bibitem{hjelm2018learning}
Hjelm, R.D., Fedorov, A., Lavoie-Marchildon, S., Grewal, K., Bachman, P.,
  Trischler, A., Bengio, Y.: Learning deep representations by mutual
  information estimation and maximization. In: International Conference on
  Learning Representations (2018)

\bibitem{ju2023comprehensive}
Ju, W., Fang, Z., Gu, Y., Liu, Z., Long, Q., Qiao, Z., Qin, Y., Shen, J., Sun,
  F., Xiao, Z., et~al.: A comprehensive survey on deep graph representation
  learning. arXiv preprint arXiv:2304.05055  (2023)

\bibitem{kipf2016semi}
Kipf, T.N., Welling, M.: Semi-supervised classification with graph
  convolutional networks. arXiv preprint arXiv:1609.02907  (2016)

\bibitem{lee2022augmentation}
Lee, N., Lee, J., Park, C.: Augmentation-free self-supervised learning on
  graphs. In: Proceedings of the AAAI Conference on Artificial Intelligence.
  vol.~36, pp. 7372--7380 (2022)

\bibitem{li2022let}
Li, S., Wang, X., Zhang, A., Wu, Y., He, X., Chua, T.S.: Let invariant
  rationale discovery inspire graph contrastive learning. In: International
  conference on machine learning. pp. 13052--13065. PMLR (2022)

\bibitem{morris2020tudataset}
Morris, C., Kriege, N.M., Bause, F., Kersting, K., Mutzel, P., Neumann, M.:
  Tudataset: A collection of benchmark datasets for learning with graphs. arXiv
  preprint arXiv:2007.08663  (2020)

\bibitem{narayanan2017graph2vec}
Narayanan, A., Chandramohan, M., Venkatesan, R., Chen, L., Liu, Y., Jaiswal,
  S.: graph2vec: Learning distributed representations of graphs. arXiv preprint
  arXiv:1707.05005  (2017)

\bibitem{oord2018representation}
Oord, A.v.d., Li, Y., Vinyals, O.: Representation learning with contrastive
  predictive coding. arXiv preprint arXiv:1807.03748  (2018)

\bibitem{popal2019guide}
Popal, H., Wang, Y., Olson, I.R.: A guide to representational similarity
  analysis for social neuroscience. Social Cognitive and Affective Neuroscience
   \textbf{14}(11),  1243--1253 (2019)

\bibitem{radford2019language}
Radford, A., Wu, J., Child, R., Luan, D., Amodei, D., Sutskever, I.: Language
  models are unsupervised multitask learners. OpenAI blog  \textbf{1}(8), ~9
  (2019)

\bibitem{ren2021label}
Ren, Y., Bai, J., Zhang, J.: Label contrastive coding based graph neural
  network for graph classification. In: Database Systems for Advanced
  Applications. pp. 123--140. Springer International Publishing (2021)

\bibitem{ren2019heterogeneous}
Ren, Y., Liu, B., Huang, C., Dai, P., Bo, L., Zhang, J.: Heterogeneous deep
  graph infomax. arXiv preprint arXiv:1911.08538  (2019)

\bibitem{ren2020adversarial}
Ren, Y., Wang, B., Zhang, J., Chang, Y.: Adversarial active learning based
  heterogeneous graph neural network for fake news detection. In: 2020 IEEE
  International Conference on Data Mining (ICDM). pp. 452--461. IEEE (2020)

\bibitem{shen2023neighbor}
Shen, X., Sun, D., Pan, S., Zhou, X., Yang, L.T.: Neighbor contrastive learning
  on learnable graph augmentation. In: Proceedings of the AAAI Conference on
  Artificial Intelligence. vol.~37, pp. 9782--9791 (2023)

\bibitem{shervashidze2011weisfeiler}
Shervashidze, N., Schweitzer, P., Van~Leeuwen, E.J., Mehlhorn, K., Borgwardt,
  K.M.: Weisfeiler-lehman graph kernels. Journal of Machine Learning Research
  \textbf{12}(9) (2011)

\bibitem{shervashidze2009efficient}
Shervashidze, N., Vishwanathan, S., Petri, T., Mehlhorn, K., Borgwardt, K.:
  Efficient graphlet kernels for large graph comparison. In: Artificial
  intelligence and statistics. pp. 488--495. PMLR (2009)

\bibitem{sun2019infograph}
Sun, F.Y., Hoffmann, J., Verma, V., Tang, J.: Infograph: Unsupervised and
  semi-supervised graph-level representation learning via mutual information
  maximization. arXiv preprint arXiv:1908.01000  (2019)

\bibitem{suresh2021adversarial}
Suresh, S., Li, P., Hao, C., Neville, J.: Adversarial graph augmentation to
  improve graph contrastive learning. Advances in Neural Information Processing
  Systems  \textbf{34},  15920--15933 (2021)

\bibitem{velivckovic2017graph}
Veli{\v{c}}kovi{\'c}, P., Cucurull, G., Casanova, A., Romero, A., Lio, P.,
  Bengio, Y.: Graph attention networks. arXiv preprint arXiv:1710.10903  (2017)

\bibitem{velivckovic2018deep}
Veli{\v{c}}kovi{\'c}, P., Fedus, W., Hamilton, W.L., Li{\`o}, P., Bengio, Y.,
  Hjelm, R.D.: Deep graph infomax. arXiv preprint arXiv:1809.10341  (2018)

\bibitem{vinyals2015order}
Vinyals, O., Bengio, S., Kudlur, M.: Order matters: Sequence to sequence for
  sets. arXiv preprint arXiv:1511.06391  (2015)

\bibitem{xiao2020should}
Xiao, T., Wang, X., Efros, A.A., Darrell, T.: What should not be contrastive in
  contrastive learning. arXiv preprint arXiv:2008.05659  (2020)

\bibitem{xu2018powerful}
Xu, K., Hu, W., Leskovec, J., Jegelka, S.: How powerful are graph neural
  networks? arXiv preprint arXiv:1810.00826  (2018)

\bibitem{yanardag2015deep}
Yanardag, P., Vishwanathan, S.: Deep graph kernels. In: Proceedings of the 21th
  ACM SIGKDD international conference on knowledge discovery and data mining.
  pp. 1365--1374 (2015)

\bibitem{ying2018hierarchical}
Ying, Z., You, J., Morris, C., Ren, X., Hamilton, W., Leskovec, J.:
  Hierarchical graph representation learning with differentiable pooling. In:
  Advances in neural information processing systems. pp. 4800--4810 (2018)

\bibitem{you2020graph}
You, Y., Chen, T., Sui, Y., Chen, T., Wang, Z., Shen, Y.: Graph contrastive
  learning with augmentations. Advances in Neural Information Processing
  Systems  \textbf{33} (2020)

\bibitem{yu2022graph}
Yu, J., Yin, H., Xia, X., Chen, T., Cui, L., Nguyen, Q.V.H.: Are graph
  augmentations necessary? simple graph contrastive learning for
  recommendation. In: Proceedings of the 45th international ACM SIGIR
  conference on research and development in information retrieval. pp.
  1294--1303 (2022)

\bibitem{zhang2018end}
Zhang, M., Cui, Z., Neumann, M., Chen, Y.: An end-to-end deep learning
  architecture for graph classification. In: Proceedings of the AAAI Conference
  on Artificial Intelligence. vol.~32 (2018)

\bibitem{zhou2022graph}
Zhou, Y., Zheng, H., Huang, X., Hao, S., Li, D., Zhao, J.: Graph neural
  networks: Taxonomy, advances, and trends. ACM Transactions on Intelligent
  Systems and Technology (TIST)  \textbf{13}(1),  1--54 (2022)

\end{thebibliography}

\end{document}